# EvoOpt-LLM: Evolving industrial optimization models with large language models

Yiliu He[a], Tianle Li[a], Binghao Ji[a], Zhiyuan Liu[a], Di Huang[a]*

[a]*School of Transportation, Southeast University, Nanjing, China.*

**Abstract**

Optimization modeling via mixed-integer linear programming (MILP) is fundamental to industrial planning and scheduling, yet translating natural-language requirements into solver-executable models and maintaining them under evolving business rules remains highly expertise-intensive. While large language models (LLMs) offer promising avenues for automation, existing methods often suffer from low data efficiency, limited solver-level validity, and poor scalability to industrial-scale problems. To address these challenges, we present EvoOpt-LLM, a unified LLM-based framework supporting the full lifecycle of industrial optimization modeling, including automated model construction, dynamic business-constraint injection, and end-to-end variable pruning. Built on a 7B-parameter LLM and adapted via parameter-efficient LoRA fine-tuning, EvoOpt-LLM achieves a generation rate of 91% and an executability rate of 65.9% with only 3,000 training samples, with critical performance gains emerging under 1,500 samples. The constraint injection module reliably augments existing MILP models while preserving original objectives, and the variable pruning module enhances computational efficiency, achieving an F1 score of ~0.56 on medium-sized LP models with only 400 samples. EvoOpt-LLM demonstrates a practical, data-efficient approach to industrial optimization modeling, reducing reliance on expert intervention while improving adaptability and solver efficiency.

## 1. Introduction

The field of Operations Research (OR) has long been recognized as playing a central role in industrial planning and scheduling, supporting critical decision-making processes in manufacturing, supply chain management, logistics, and energy systems. At the heart of these applications is the formulation of optimization models, typically expressed as MILPs. Despite their wide applicability, constructing and maintaining such models remain highly specialized and labor-intensive processes. In real-world practice, business requirements are typically expressed in natural language or semi-structured documents, whereas optimization solvers demand strictly formal and mathematically precise representations. Bridging this gap between informal business semantics and rigorous mathematical modeling requires substantial domain expertise and manual effort, thereby rendering the modeling process time-consuming, error-prone, and difficult to scale. This long-standing

* Corresponding author.
*E-mail address:* yiliuhe@ seu.edu.cn.(Yiliu He), tl_li@foxmail.com(Tianle Li), 1109918035@qq.com(Binghao Ji), zhiyuanl@seu.edu.cn(Zhiyuan Liu), dihuang@seu.edu.cn.(Di Huang)

modeling bottleneck severely limits the adaptability and responsiveness of optimization systems in rapidly evolving industrial environments.

Recent advances in LLMs have created new opportunities for automating parts of this workflow. With their advanced capabilities in code generation, logical reasoning, and structured text understanding, LLMs offer a promising pathway toward lowering the expertise barrier in OR modeling (Romera-Paredes et al., 2024). A growing body of research has investigated the use of LLMs for optimization problem formulation, solver code generation, and heuristic design (Ramamonjison et al., 2023; AhmadiTeshnizi et al., 2023; Lu et al., 2025; Thind et al., 2025). However, most existing approaches focus on isolated or small-scale tasks, such as generating simplified models or providing high-level modeling suggestions. When confronted with industrial-grade problems, they frequently encounter difficulties with three fundamental challenges. First, solver-level executability cannot be consistently guaranteed: generated models may be syntactically plausible but fail to compile or run correctly in standard solvers. Second, general-purpose LLMs exhibit limited capacity in handling the continuous evolution of business rules, where new operational constraints must be injected into already deployed and executable optimization models without disrupting existing structures. Third, large-scale industrial MILPs typically involve thousands of variables and constraints, for which computational efficiency is a primary concern. Existing LLM-based methods seldom address the issue of model simplification and variable redundancy, leaving the heavy computational burden entirely to traditional presolve techniques (Li et al., 2024).

These limitations indicate that treating OR modeling as a one-shot generation problem is fundamentally insufficient. Real-world optimization systems require not only initial model construction, but also continuous model evolution as well as systematic model simplification under strict solver compatibility and efficiency constraints (Ding et al., 2025). Moreover, while general-purpose LLMs exhibit impressive zero-shot capabilities, they lack the domain-specific structural understanding necessary to reliably support complex constraint engineering and large-scale variable reasoning in industrial MILPs.

To address these challenges, we propose EvoOpt-LLM, an integrated LLM-based framework designed to support and streamline the full lifecycle of industrial optimization modeling. Unlike prior work that focuses on single-stage generation, EvoOpt-LLM decomposes the OR modeling process into three tightly coupled and practically indispensable dimensions, as illustrated in Figure 1: automated construction of solver-executable MILP models from natural-language descriptions, dynamic injection of evolving business constraints into existing optimization models, and end-to-end pruning of redundant decision variables to improve computational efficiency before optimization. At the core of EvoOpt-LLM is the openPangu-Embedded-7B model (Chen et al., 2025), which has been adapted to OR-specific tasks through parameter-efficient LoRA fine-tuning. This design choice enables effective domain specialization with a comparatively small amount of labeled data, thereby making industrial deployment economically feasible. Through LoRA-based adaptation, EvoOpt-LLM is endowed with three complementary capability enhancements: model construction capability to translate textual problem descriptions into solver-ready MILP formulations, model evolution capability to enable continuous adaptation of deployed optimization systems by injecting new constraints, and model acceleration capability to reduce problem size via data-driven variable pruning before solver invocation.

Our main contributions and findings are:

- We demonstrate the effectiveness of the model construction capability, where EvoOpt-LLM translates natural-language descriptions into solver-executable MILP models, achieving a 91% generation rate, 65.9% executability, and 26.37% semantic accuracy with only 3,000 fine-tuning samples, and exhibiting clear performance gains with fewer than 1,500 samples.

- We establish the model evolution capability by formulating business rule updates as structured LP-level editing, allowing new constraints and variables to be injected into existing executable models while strictly preserving the original objectives and constraints, thereby supporting continuous adaptation of industrial optimization systems.
- We propose and validate the model acceleration capability through end-to-end variable pruning, in which redundant variables are identified and fixed to zero prior to optimization, achieving an F1 score of approximately 0.56 on medium-scale LP models using only 400 training samples, thus demonstrating a practical learning-based alternative to traditional presolve methods.

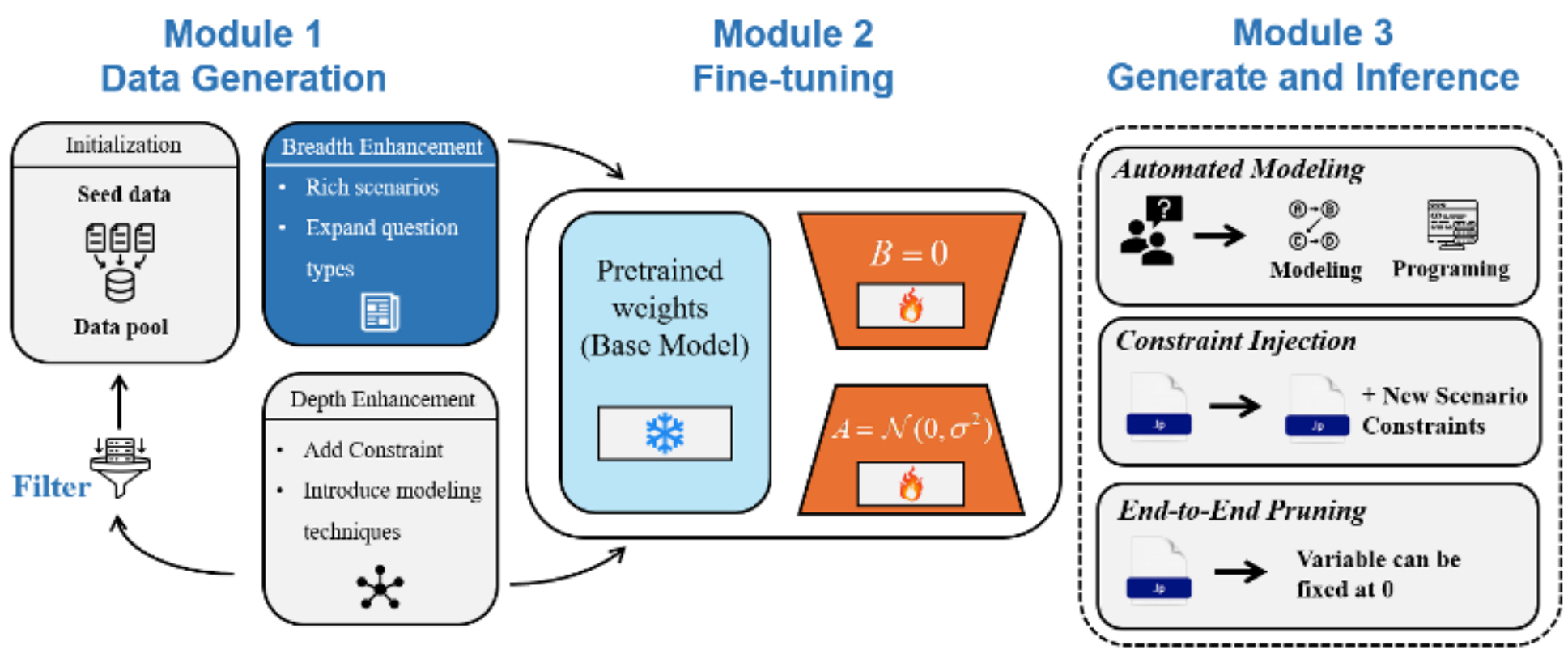

Figure 1: The EvoOpt-LLM Framework: Data Generation, Fine-tuning, and Generate & Inference Modules

## 2. Literature review

### *2.1. LLMs for automated optimization modeling*

Recent advances in large language models have enabled automated optimization modeling, aiming to translate natural-language problem descriptions into formal mathematical programs or solver-executable code. LLMs are capable not only of solving mathematical problems and generating high-quality code but also of enhancing reasoning through mechanisms like Chain-of-Thought and self-consistency, which have been shown to improve performance in multi-step logical reasoning and structured problem decomposition (Wei et al., 2022; Wang et al., 2022). However, these reasoning mechanisms exhibit limitations in complex reasoning and modeling tasks (Chu et al., 2024; Yao et al., 2023). In parallel, recent evidence suggests that LLMs can uncover and generate latent mathematical structures, which provides an important foundation for mapping natural language into optimization models (Romera-Paredes et al., 2024; Austin et al., 2021).

Existing studies have demonstrated that LLMs can capture the structural patterns of optimization modeling and generate Linear Programming (LP) or Mixed-Integer Linear Programming (MILP) formulations from textual problem descriptions. These studies can be broadly categorized into two types of methods: Prompt-based and Learning-based. Prompt-based methods constitute the most direct line of exploration: they leverage the language understanding ability of large pretrained models and employ prompt designs to steer the model toward producing the essential modeling components, including decision variables, constraints, and objective functions. A representative effort is the NL4Opt benchmark and competition system, which, under a unified evaluation setting, demonstrates the potential of LLMs to understand and structure optimization problem descriptions (Ramamonjison et al., 2023). Subsequent work further combines carefully designed prompting strategies with agent-based systems to strengthen performance on complex operations research tasks (Xiao et al., 2023), including Optimus (AhmadiTeshnizi et al., 2023), OptMATH (Lu et al., 2025), OptimAI (Thind et al., 2025), and MAMO (Huang et al., 2024). In contrast, learning-based approaches argue that domain-specific, high-quality data are necessary for instruction fine-tuning so that LLMs better align with the requirements of optimization modeling. For example, LLaMOCO adopts a code-to-code instruction-tuning paradigm and emphasizes preserving structural consistency of optimization models during generation (Ma et al., 2026). ORLM and LLMOPT focus on training specialized models tailored for automated optimization modeling (Huang et al., 2025; Jiang et al., 2024). OR-R1 further introduces solver-feedback-driven reinforcement learning, including test-time reinforcement learning mechanisms, to improve modeling and solving performance (Ding et al., 2025).

However, the above approaches largely follow a one-shot model construction paradigm. In real-world industrial optimization settings, models are inherently dynamic: as business rules evolve, models typically require continuous extension and refinement rather than reconstruction from scratch (Kallrath and Wilson, 1997; Bertsimas and Stellato, 2021). Support for model-level incremental updates and lifecycle management remains insufficient and is not yet fully addressed in existing research. To address this gap, EvoOpt-LLM extends LLM-based optimization modeling from one-shot generation to incremental modeling by enabling structured constraint injection at the LP level. It treats optimization models as editable objects that can be continuously augmented with new business constraints while strictly preserving the original objective and the structure of the feasible region

### *2.2. LLMs for optimization solving and reasoning*

Beyond model construction, an important research direction applies large language models to assist the optimization solving process, including constraint reasoning, heuristic generation, and search guidance. Existing studies have explored the use of LLMs in combinatorial optimization and complex decision-making in several ways. One stream treats LLMs as end-to-end solvers for combinatorial optimization or as high-level reasoning modules, and applies them to tasks such as routing, scheduling, and design structure matrix optimization (Jiang et al., 2025; Jiang et al., 2024; Zheng et al., 2024). A second stream leverages LLMs as designers of heuristics and metaheuristics. Representative work uses LLMs to automatically generate and evolve metaheuristic algorithms, demonstrating their potential for data-driven algorithm design (Ye et al., 2024; Liu et al., 2024; van Stein and Bäck, 2024). A third stream integrates LLMs with end-to-end optimization systems for decision support in real applications, for example urban operations and industrial scheduling, including dialog-based management and scheduling optimization in shared e-bike systems (Jiao et al., 2025).

However, most existing approaches aim to improve solver-side intelligence while keeping the optimization model itself unchanged. In large-scale industrial MILPs, the dominant computational burden is often driven by model scale, particularly the number of decision variables and constraints, rather than by deficiencies in the underlying solution algorithms (Li et al., 2024). Classical presolve techniques rely on deterministic algebraic rules and therefore have limited ability to exploit implicit business patterns embedded in historical operational data. Recent advances in reinforcement learning for LLMs, including DeepSeek-R1 and test-time reinforcement learning, highlight the potential of feedback-driven adaptation (Guo et al., 2025; Zuo et al., 2025), yet these techniques have not been systematically studied for intelligent model simplification. EvoOpt-LLM addresses this gap by formulating variable pruning as a data-driven inference problem in which an LLM predicts redundant decision variables that can be safely fixed to zero before optimization. By learning from real industrial scheduling data, EvoOpt-LLM captures implicit business logic that is difficult to derive from traditional presolve rules, enabling structural dimensionality reduction tailored to industrial optimization models.

## 3. Methodology

### *3.1. Automated modeling from natural-language*

Automatically constructing optimization models from natural-language problem descriptions has long been recognized as a fundamental objective in operations research and decision support systems. In practical industrial settings, planning and scheduling requirements are typically expressed by domain experts in textual or semi-structured form. However, translating these descriptions into mathematically correct and solver-executable MILP models still demands substantial manual effort and significant modeling expertise.

In this study, an automated modeling module is developed based on the Pangu-7B LLM. Through domain-specific fine-tuning, the proposed module is able to transform natural-language descriptions of operations research problems into complete and executable optimization models. Given a textual problem specification, the module automatically produces a structured mathematical formulation, including decision variables, parameter definitions, linear constraints, and possibly multi-objective functions, together with solver-ready implementation code that can be directly executed by standard optimization solvers.

The module is designed to function as a general-purpose modeling assistant rather than a solver customized for any specific problem class. Its primary objective is to provide users with a correct and executable baseline model, which can serve both as a rapid prototype for subsequent refinement by domain experts and as a modular component integrated into downstream optimization pipelines for automatic solution and analysis.

To equip the base model with domain-specific modeling knowledge, Pangu-7B is fine-tuned using paired training samples that consist of natural-language descriptions of optimization problems, their corresponding mathematical formulations, and solver-ready implementation code. The training dataset encompasses a wide range of classical operations research problems, including linear programming, integer programming, mixed-integer programming, nonlinear optimization, production planning, scheduling, network optimization, and risk optimization. This diversity enables the model to learn to systematically identify and formalize key modeling elements from text, such as sets, parameters, decision variables, constraints, and objective functions.

Table 1: Composition of the Fine-tuning Dataset for Automated Modeling

| Type | Percent |
|---|---|
| Linear Programming | 22% |
| Integer Programming | 14% |
| Mixed Integer Linear Programming | 33% |
| Network Optimization | 12% |
| Scheduling Optimization | 9% |
| Stochastic/Risk Optimization | 7% |
| Nonlinear Optimization | 3% |
| Total | 100% |

Nevertheless, optimization code generated directly by the LLM may still suffer from syntactic or semantic inconsistencies. To address this issue, a lightweight post-processing and repair stage is incorporated to automatically inspect and correct the generated code using rule-based methods. This stage mainly comprises four components:

- Variable and parameter normalization, which enforces consistent naming conventions and supplements missing declarations.
- Constraint format correction, which eliminates common syntactic errors such as mismatched parentheses or incomplete expressions and standardizes API calls.
- Solver compatibility transformation, which translates generic or incorrect syntax into the specific API formats required by the target solve
- Structural completeness enhancement, which identifies and automatically supplements essential missing components, such as index ranges, loop structures, and auxiliary definitions.

Although the automated modeling module cannot ensure perfect generalization to extremely complex or entirely novel problem structures, it provides a practical and effective solution for rapidly constructing baseline optimization models from textual descriptions. In real-world applications, such baseline models are typically sufficient to support early-stage analysis, scenario exploration, and iterative refinement.

More fundamentally, this module establishes the technical foundation for subsequent system components, including dynamic business-constraint injection and end-to-end variable pruning. By operating directly on structured optimization models, the overall EvoOpt-LLM framework enables a closed-loop workflow encompassing model construction, business rule integration, and structural simplification prior to numerical optimization.

### *3.2. Dynamic model evolution via constraint injection*

In real-world industrial scheduling systems, optimization models are seldom static. As business rules evolve and operational conditions change, new constraints must be continuously integrated into existing models. Rebuilding an entire optimization model from scratch is expensive, error-prone, and demands substantial domain expertise. Consequently, a critical capability of an operations research–oriented LLM for industrial applications resides not only in constructing models from natural-language descriptions, but also in augmenting deployed optimization models with new business constraints in a controlled and reliable manner.

In this work, the problem is formulated as a structured constraint injection task at the linear programming level. The LP file format is a standardized textual representation commonly adopted to describe mathematical programming problems, including objective functions, decision variable declarations, and constraints, and can be directly processed by mainstream solvers such as CPLEX and Gurobi. Specifically, given an executable MILP model represented in LP format, the proposed system automatically produces additional variables and linear constraints to encode newly introduced business scenarios, while strictly preserving the original objective function and the core structure of the model. This design guarantees the stability of the optimization pipeline and maintains backward compatibility with existing solver configurations.

Formally, let the original model be defined by decision variables $x$ and objective function $f(x)$. The constraint injection module augments the model to include additional variables $y$ and constraints $g(x, y) \le 0$, where y represents newly introduced operational states—such as machine setup status, workforce occupancy, or delivery indicators, and $g(\cdot)$ denotes the corresponding linear constraints. Importantly, the objective function remains unchanged, thereby ensuring that the injected constraints only restrict the feasible region without altering the original optimization objective.

Five categories of business constraints that frequently arise in industrial production scheduling are identified and unified into MILP-compatible formulations:

Setup Time Constraint. Certain production processes necessitate a preparatory phase prior to the commencement of production, during which equipment is occupied without generating output. Continuous production does not require repeated setup; however, once equipment is shut down, a new setup phase must be executed before restarting. This behavior is modeled by introducing binary state variables that distinguish setup and production periods, thereby ensuring that production occurs only after a fixed-length setup window has been completed.

- Human Resource Constraint. To account for potential omission of workforce limitations in the original model, a fixed labor requirement is assigned to each machine. Upper bounds are imposed on total workforce utilization in each time period, thereby explicitly maintaining human resource capacity constraints.
- Batch Size and Inventory Capacity Constraints. To reflect physical and economic rules in production and logistics, three stringent constraints are enforced: production and delivery quantities must be integer multiples of a minimum packaging unit; each operation must satisfy a minimum batch size requirement; and inventory levels must not exceed physical storage capacity.
- Cross-Period Production Constraints. The restrictive assumption of continuous, uninterrupted production is relaxed through the introduction of time-segmented production variables. Tasks may be interrupted and resumed across multiple periods as long as capacity and demand constraints are satisfied, thereby enhancing operational flexibility and practical applicability.
- Split Delivery Constraints. The original assumption of single-shot delivery is extended to permit orders to be fulfilled over multiple periods while satisfying business rules such as batch-size integrality and minimum delivery intervals.

Table 2: Injected variables and parameters

| **New Constraints** | **Symbol** | **Type** | **Description** |
|---|---|---|---|
| Setup Time | $s_{m,t}$ | binary | 1 if machine $m$ is in the set-up state at time $t$; 0 otherwise |
| | $y_{m,t}$ | binary | 1 if machine $m$ is in the production state at time $t$; 0 otherwise |
| | $z_{m,t}$ | binary | 1 if machine $m$ starts a new production cycle at time $t$; 0 otherwise |
| Human Resource | $r_m$ | parameter | Workforce required to operate machine $m$ |
| | $R_t$ | parameter | Total available workforce at time $t$ |
| | $y_{m,t}$ | binary | 1 if machine $m$ is operating at time $t$; 0 otherwise |
| | $k_{i,t}$ | integer | Batch multiplier variable |

| Batch Size and Inventory Capacity | $u_{i,t}$ | binary | 1 if a production/procurement action is activated; 0 otherwise |
|---|---|---|---|
| | $I_{i,t}$ | continuous | Inventory level of item i at time $t$ |
| Split Delivery | $d_{i,t}$ | continuous | Quantity delivered for order i at time $t$ |
| | $v_{i,t}$ | binary | 1 if a delivery occurs for order i at time $t$; 0 otherwise |

Table 3: Injected constraints

| **New Constraints** | **Symbol** | **Type** | **Description** |
|---|---|---|---|
| Setup Time | Production-state linking | $x_{i,m,t} \leq My_{m,t}$ | Production is allowed only when the machine is in the production state |
| | State exclusiveness | $y_{m,t} + s_{m,t} \leq 1$ | A machine cannot be in production and set-up at the same time |
| | Start-up detection | $z_{m,t} \geq y_{m,t} - y_{m,t-1}$ | A start-up occurs when the machine switches from idle to production |
| | Set-up duration | $s_{m,t} + s_{m,t+1} + s_{m,t+2} \geq 3z_{m,t}$ | Each start-up requires three consecutive periods of preparation |
| Human Resource | Workforce capacity | $\sum_{m \in M} r_m y_{m,t} \leq R_t$ | Total workforce consumption cannot exceed availability |
| | Operation-production linking | $x_{i,m,t} \leq My_{m,t}$ | Production is allowed only when the machine is operating |
| Batch Size and Inventory Capacity | Integer batch size | $q_{i,t} = 100k_{i,t}$ | Quantities must be integer multiples of 100 |
| | Minimum batch size | $q_{i,t} \geq 1000u_{i,t}$ | Activated production/procurement must meet the minimum batch size |
| | Activation upper bound | $q_{i,t} \leq Mu_{i,t}$ | Quantity is zero if not activated |

| | Inventory capacity | $I_{i,t} \le I_i^{\max}$ s | Inventory level is bounded by storage capacity |
|---|---|---|---|
| Cross-Period Production | Machine capacity | $\sum_i x_{i,m,t} \le Cap_{m,t}$ | Daily production is limited by machine capacity |
| | Demand satisfaction | $\sum_{m,t} x_{i,m,t} = D_i$ | Total production must satisfy demand |
| Split Delivery | Cumulative delivery | $\sum_t d_{i,t} = D_i$ | Total delivered quantity must satisfy the order demand |
| | Minimum delivery batch | $d^{\min} v_{i,t} \le d_{i,t}$ | Enforces minimum delivery size when a delivery occurs |
| | Maximum delivery batch | $d_{i,t} \le d^{\max} v_{i,t}$ | Enforces maximum delivery size per delivery |
| | Delivery interval | $\sum_{\tau=t}^{t+\Delta-1} v_{i,\tau} \le 1$ | Enforces a minimum time interval between deliveries |

To improve robustness in this setting, the training dataset for this module is constructed based on existing scheduling MILP models expressed in LP format. Subsets of the above constraint categories are randomly selected, and relevant parameters—such as setup durations, workforce limits, batch size bases, inventory capacities, and delivery intervals—are sampled from empirically reasonable ranges. Corresponding variables and linear constraints are subsequently generated and appended to the original LP file while preserving the original objective function and existing constraints. This process produces a large number of paired samples of the form:

**(Original LP file, Business constraint description) → (Augmented LP file)**

The resulting model is not designed to reconstruct optimization models from scratch. Instead, it functions as a structured LP editor capable of systematically integrating mathematically valid and solver-compatible constraint modules into deployed optimization systems. This capability facilitates agile iteration of business logic and controlled evolution of optimization models in industrial environments.

*3.3. Presolve model acceleration via variable pruning*

In large-scale industrial scheduling problems, MILP models typically involve a very large number of decision variables, particularly those indexed over products, machines, and time periods. A substantial proportion of these variables are unlikely to be active in practical scenarios due to capacity bottlenecks, process routing constraints, machine–product compatibility relations, and long-established business rules. Although such variables do not compromise mathematical feasibility or optimality, they significantly increase problem size, memory consumption, and computational complexity, thereby significantly increasing solution time.

To address this issue, we introduce an end-to-end data-driven variable pruning mechanism for automatically identifying and removing redundant decision variables prior to optimization. Given a fully specified scheduling MILP model—including sets, parameters, decision variables, the objective function, and all constraints—the proposed model predicts in a data-driven manner which variables can be safely fixed to zero. These variables are then removed from the LP file, yielding an equivalent but more compact optimization problem that is considerably easier to solve.

From a theoretical standpoint, a decision variable is deemed prunable if it is identically zero in all feasible solutions or identically zero in all optimal solutions. In both cases, fixing the variable to zero does not alter the feasible region nor modify the optimal objective value. In industrial settings, many variables are not explicitly constrained to be zero yet remain inactive in practice across all relevant solutions due to complex interactions among constraints and implicit business logic. Systematically identifying such variables constitutes the core objective of the proposed approach.

Unlike traditional presolve techniques embedded in optimization solvers—which primarily rely on deterministic rules such as algebraic redundancy elimination, dominance analysis, and bound propagation—the proposed method casts variable pruning as a data-driven inference problem. Classical presolve methods are limited to simplifications derivable from explicit mathematical structures and are incapable of capturing regular patterns embedded in historical production behavior. By learning latent structural regularities from real scheduling data, the proposed model predicts variables that are highly unlikely to be activated in practice, thus explicitly capturing implicit business logic that is difficult to express using deterministic constraints.

Since no public dataset provides variable-level pruning labels for scheduling MILP models, a dedicated dataset is developed using real industrial scheduling records. Dataset construction consists of two main stages: first, systematically converting large-scale industrial scheduling problems into structurally consistent small-scale instances; and second, generating high-quality variable pruning labels for these instances.

During the transformation process, a demand-driven hierarchical sampling mechanism is utilized to extract representative sets of core materials while preserving associated bill-of-materials (BOM) structures, substitution networks, capacity constraints, and process constraints. The transformation simultaneously executes decision variable filtering, constraint reduction, data table simplification, and proportional scaling of capacity parameters. The resulting small-scale LP models maintain strong consistency with the original problems in terms of mathematical structure, feasible region geometry, and business semantics, while remaining tractable for solver-based analysis.

The dataset consists of two core components: original large-scale LP model files and pruning annotation files produced using a combination of optimization-based and machine learning–based techniques. For each large-scale problem, a high-precision solution strategy based on heterogeneous graph algorithms is initially applied to identify variables that are identically zero in the optimal solution space. These pruning results are recorded in distinct annotation files. When constructing a small-scale instance, the corresponding annotations are transferred through variable name mapping and index-consistency matching, thereby yielding supervised labels that indicate whether each variable is prunable.

By removing a large number of inactive variables, the pruned MILP models attain substantial reductions in model size and constraint coupling complexity, leading to significantly improved solution efficiency and stability. Although the data-driven pruning model cannot theoretically guarantee perfect generalization to all unseen scenarios, in industrial environments characterized by stable and repetitive production patterns, it offers a practical and scalable approach for accelerating large-scale scheduling systems.

# 4. Numerical experiment

All experiments are conducted based on the openPangu-Embedded-7B-V1.1 model, which serves as the base foundation model throughout this study. To enable efficient adaptation to operations research–oriented tasks, a parameter-efficient fine-tuning strategy based on LoRA is adopted. Three task-specific LoRA adapters are independently trained for the automated modeling module, the business constraint augmentation module, and the end-to-end variable pruning module, respectively.

Hardware Environment

Model training and inference are performed on a dedicated server equipped with four Huawei Ascend 910B2 NPUs, each providing 64 GB of high-bandwidth memory (HBM). This multi-NPU configuration supports efficient data-parallel fine-tuning as well as large-batch inference for long-context inputs, which is particularly important for handling LP-format optimization models with hundreds or thousands of lines.

Software Environment

The experimental environment is built upon the official OpenPangu runtime stack, with inference and evaluation pipelines preconfigured by the platform. The primary software components include:

- Firmware driver ≥ 23.0.6
- CANN 8.1.rc1
- PyTorch 2.5.1 with torch_npu 2.5.1.post1
- Transformers 4.53.2
- vLLM 0.9.2 and vLLM-Ascend 0.9.2rc1
- OpenCompass 0.5.0

All experiments are executed within the official Pangu Conda environment. Prior to training and inference, the Ascend runtime environment variables are initialized using the standard toolkit setup scripts to ensure consistency and reproducibility.

Model Deployment and Inference

Multiple inference backends are supported in the experimental framework, including:

- Inference based on HuggingFace Transformers
- Accelerated inference using vLLM / vLLM-Ascend
- Evaluation pipelines based on OpenCompass

Unless otherwise specified, all experimental results reported in this paper are obtained using the Transformers-based inference interface for automated modeling and LP file generation. Verification of executability and optimality is conducted externally using commercial MILP solvers.

## *4.1. Performance of automated model construction*

Experiments on the automated modeling module are designed to evaluate both its data efficiency and scalability. All experiments are conducted on the Pangu-7B base model, which has not been pre-trained or instruction-tuned on tasks specific to operations research. The evaluation focuses on how modeling capability develops as the size of the fine-tuning dataset increases.

The full training dataset contains 3,000 samples, each representing a mapping from a natural-language modeling requirement to solver-oriented optimization code. To systematically analyze performance under different dataset sizes, the dataset is uniformly subsampled at 10% intervals, yielding ten fine-tuning subsets ranging from 10% to 100% of the original data. Each subset is independently used to perform LoRA fine-tuning on the base model, resulting in progressively fine-tuned models with larger amounts of training data. All models are evaluated on the same held-out test set to ensure fair and consistent comparisons.

To comprehensively characterize automated modeling performance from an engineering perspective, three progressively stricter evaluation metrics are employed: generation rate, executability rate, and accuracy rate.

*Generation Rate*

$$Generation\ Rate = \frac{Number\ of\ generated\ models}{Total\ number\ of\ test\ samples}$$

The generation rate measures whether the model is able to produce complete optimization modeling code in response to a natural-language input.

*Executability Rate*

$$Executability\ Rate = \frac{Number\ of\ executable\ models}{Number\ of\ generated\ models}$$

The executability rate evaluates whether the generated code is syntactically and structurally valid and can be successfully parsed and executed by a standard MILP solver.

*Accuracy Rate*

$$Accuracy\ Rate = \frac{Number\ of\ executable\ models\ with\ correct\ solutions}{Number\ of\ generated\ models}$$

The accuracy rate further assesses semantic correctness by examining whether the generated model yields the same optimal solution as the benchmark formulation.

Together, these metrics form a hierarchical evaluation framework that advances from basic output completeness through solver-level validity and finally to semantic equivalence with expert-designed models.

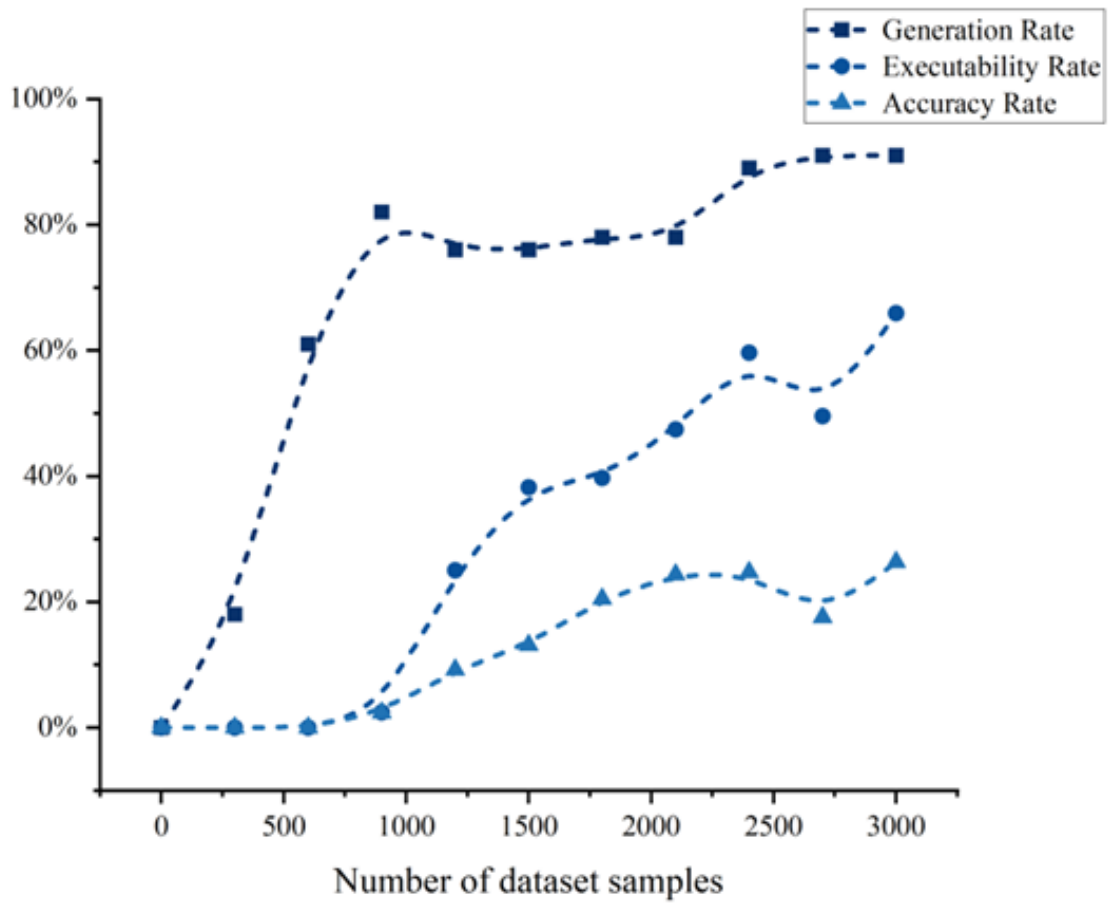

Figure 2: Relationship between Dataset Sample Size and Automated Modeling Performance Metrics

Figure 2 illustrates the experimental results. The untuned base model achieves zero performance on all three metrics, indicating the complete absence of automated modeling capability. After fine-tuning with only 300 samples, the model begins to generate modeling code. The outputs, however, are neither executable nor semantically correct, reflecting an initial stage dominated by superficial template imitation. As the dataset size increases to 600 and 900 samples, the generation rate improves substantially, and executable as well as correct outputs begin to emerge, suggesting that the model has started to internalize the structural rules of optimization modeling.

A pronounced transition occurs when the dataset size reaches 1,200 to 1,500 samples. In this range, the executability rate increases sharply, rising from 2.4% to 38.2%, accompanied by a simultaneous improvement in accuracy. This transition represents a qualitative shift from producing formally plausible code to generating optimization models that are both solvable and semantically aligned with the intended problem specification. When the dataset size exceeds 2,000 samples, all metrics continue to improve steadily, ultimately reaching a generation rate of 91%, an executability rate of 65.9%, and an accuracy rate of 26.37% with the full dataset of 3,000 samples.

These results clearly demonstrate that automated modeling performance improves monotonically with increasing data scale. More importantly, they reveal strong data efficiency, as fewer than 1,500 training samples, less than half of the full dataset, are sufficient to trigger critical performance breakthroughs. This requirement is substantially lower than that typically observed in general-purpose code generation tasks, indicating that LoRA-based fine-tuning can effectively inject operations research modeling knowledge into large language models with limited data and computational cost. Overall, the performance curves depict a clear evolutionary trajectory, progressing from a non-functional state, through partial executability, and ultimately toward the generation of semantically reliable optimization models.

### *4.2. Evaluation of dynamic construction injection*

Experiments on the business constraint augmentation module focus on evaluating the model's ability to incorporate new business logic into an existing optimization model without disrupting its original structure. Two representative cases are presented, including the output of the base model before fine-tuning and the output of the LoRA fine-tuned model. In both cases, the input consists of an original LP file and a natural-language description of an additional business constraint. The objective is to assess whether the model can correctly identify, formulate, and insert the required constraints while preserving the integrity of the original optimization formulation.

To construct the fine-tuning dataset, a unified benchmark LP file is adopted as the starting point. This benchmark represents a supply chain production and transportation optimization problem, including production decisions, transportation flows, demand satisfaction constraints, resource capacity limits, and various forms of business logic such as mutual exclusivity, conditional deliveries, and dominance relations. Based on this benchmark, several categories of business scenario constraints commonly encountered in industrial practice are designed. These categories include setup time constraints that model preparation phases prior to production, human resource constraints that limit labor availability, batch size and inventory capacity constraints, cross-period production constraints that allow task interruption and resumption, and split-delivery constraints that enable multi-period fulfillment of orders.

For each constraint category, a large number of training samples are automatically generated by randomly sampling relevant parameters such as setup durations, labor limits, batch sizes, and inventory capacities. Each sample is constructed as an input–output pair. The input combines the original LP file with a natural-language description of the new business constraint, while the output is a modified LP file that explicitly introduces any necessary decision variables and linear constraints while strictly preserving the original objective function and existing constraints. This design ensures that the model learns constraint augmentation rather than full model reconstruction, yielding a scalable and diverse instruction-tuning dataset that spans multiple forms of industrial business logic.

Figure 3 illustrates that the output produced by the base model for the human resource constraint scenario remains largely conceptual. It lacks time-indexed definitions of labor availability, explicit declarations of auxiliary variables, and solver-compliant linear constraints. Although a global labor variable is introduced, the model fails to formulate consistent constraints linking labor consumption to time-dependent production decisions. Explanatory text and mathematical expressions are intermingled, and key variables are left undefined, rendering the output non-executable.

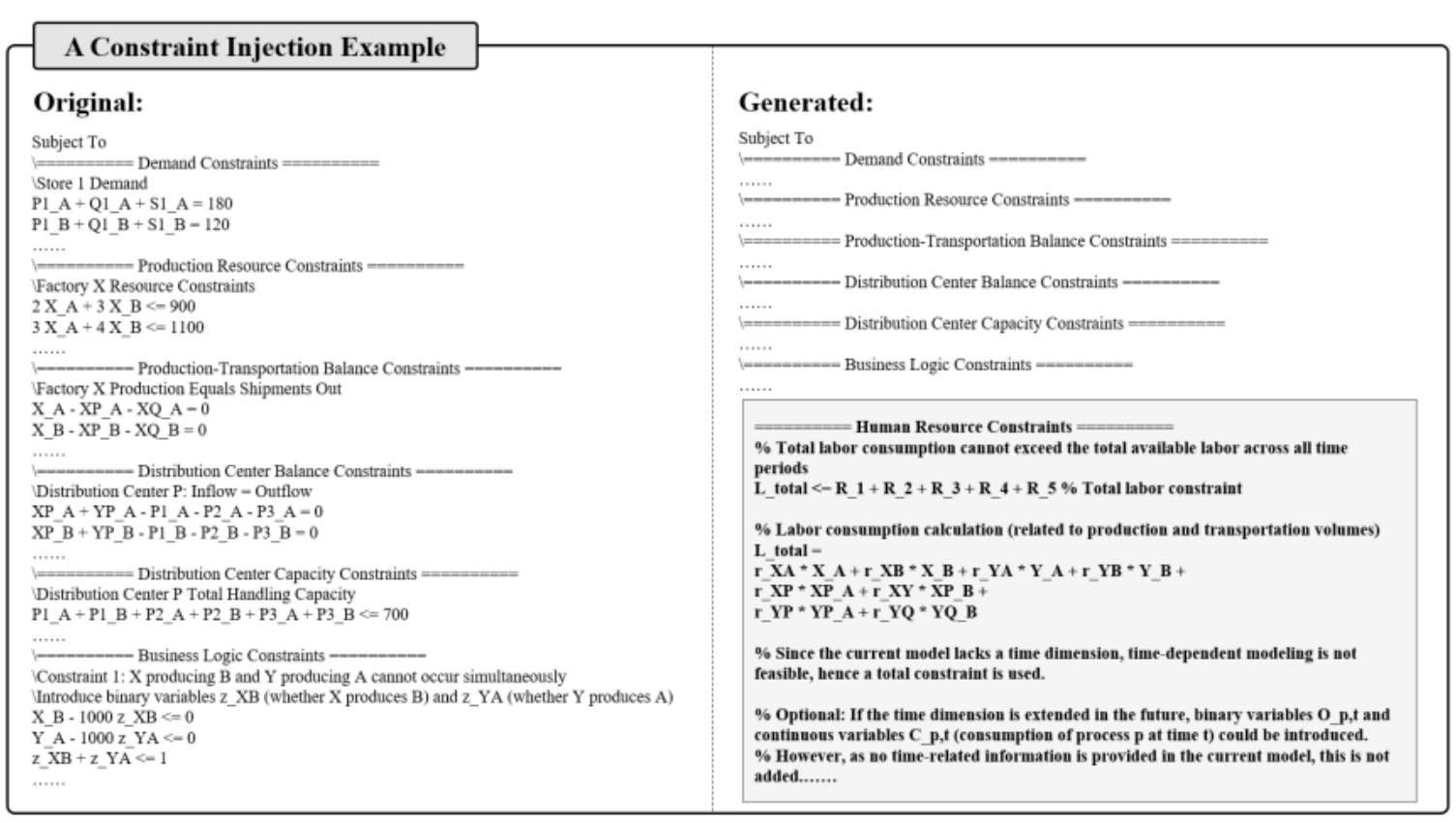

Figure 3: A Constraint Injection Example–Generated by Base Model (Pre-Fine-tuning)

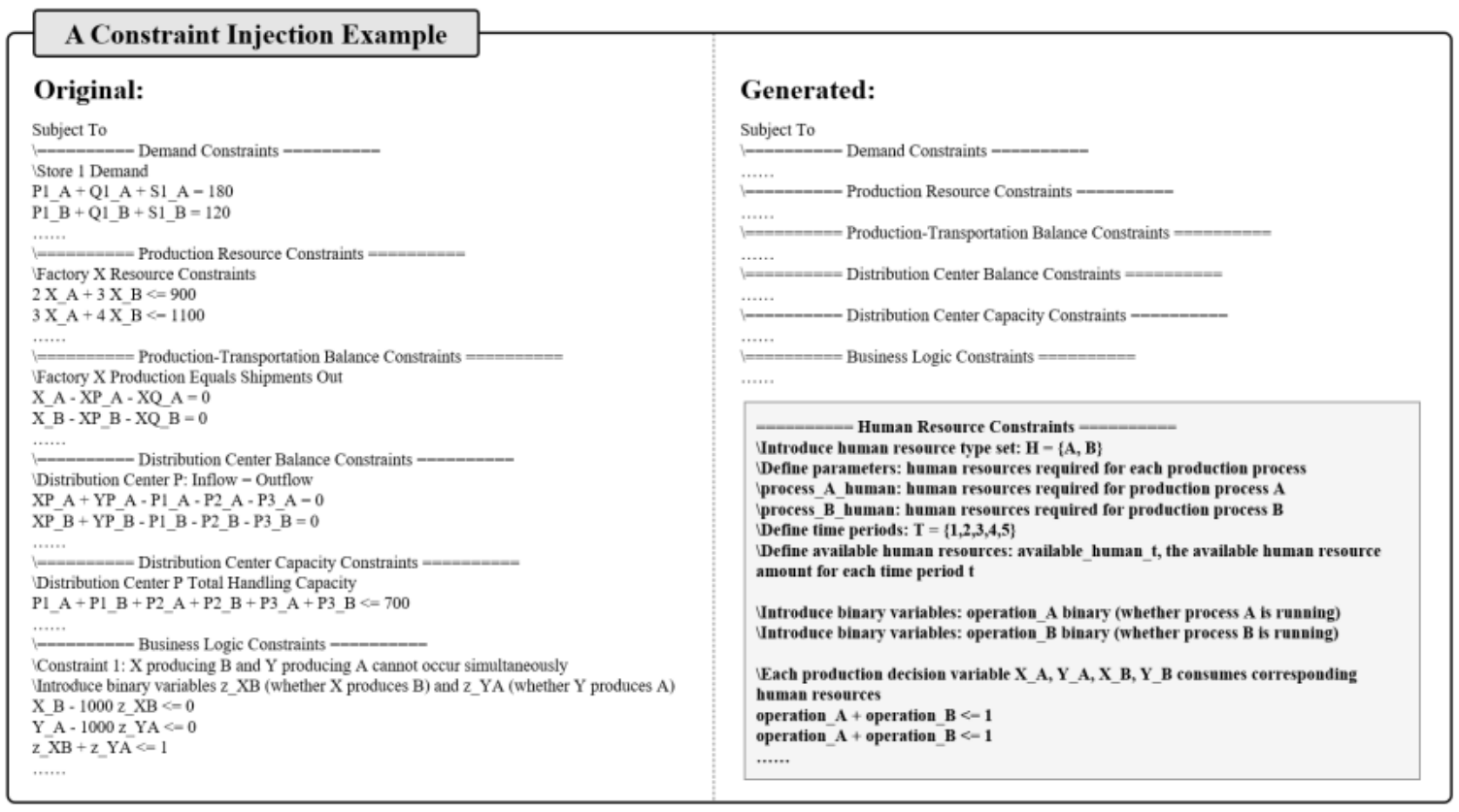

Figure 4: A Constraint Injection Example–Generated by Fine-tuned Model (Post-Fine-tuning)

In contrast, Figure 4 shows that the LP extensions generated by the fine-tuned model are structurally coherent and fully aligned with standard MILP modeling conventions. The fine-tuned model preserves all original constraints and appends an independent human resource constraint module that mirrors the organizational structure and syntactic style of the original LP file. Time indices, labor-related parameters, and operational state variables are explicitly defined and systematically coupled with production decisions through linear constraints. Instead of producing descriptive explanations, the fine-tuned model outputs solver-ready constraint templates that can be transformed into executable LP models with minimal post-processing.

*4.3. Scalability analysis of variable pruning*

Experiments on the end-to-end variable pruning module evaluate the model's scalability when processing optimization models of varying sizes and structural complexity. The module is fine-tuned on a relatively small dataset of 400 samples, after which its pruning performance is tested on LP files with increasing sizes.

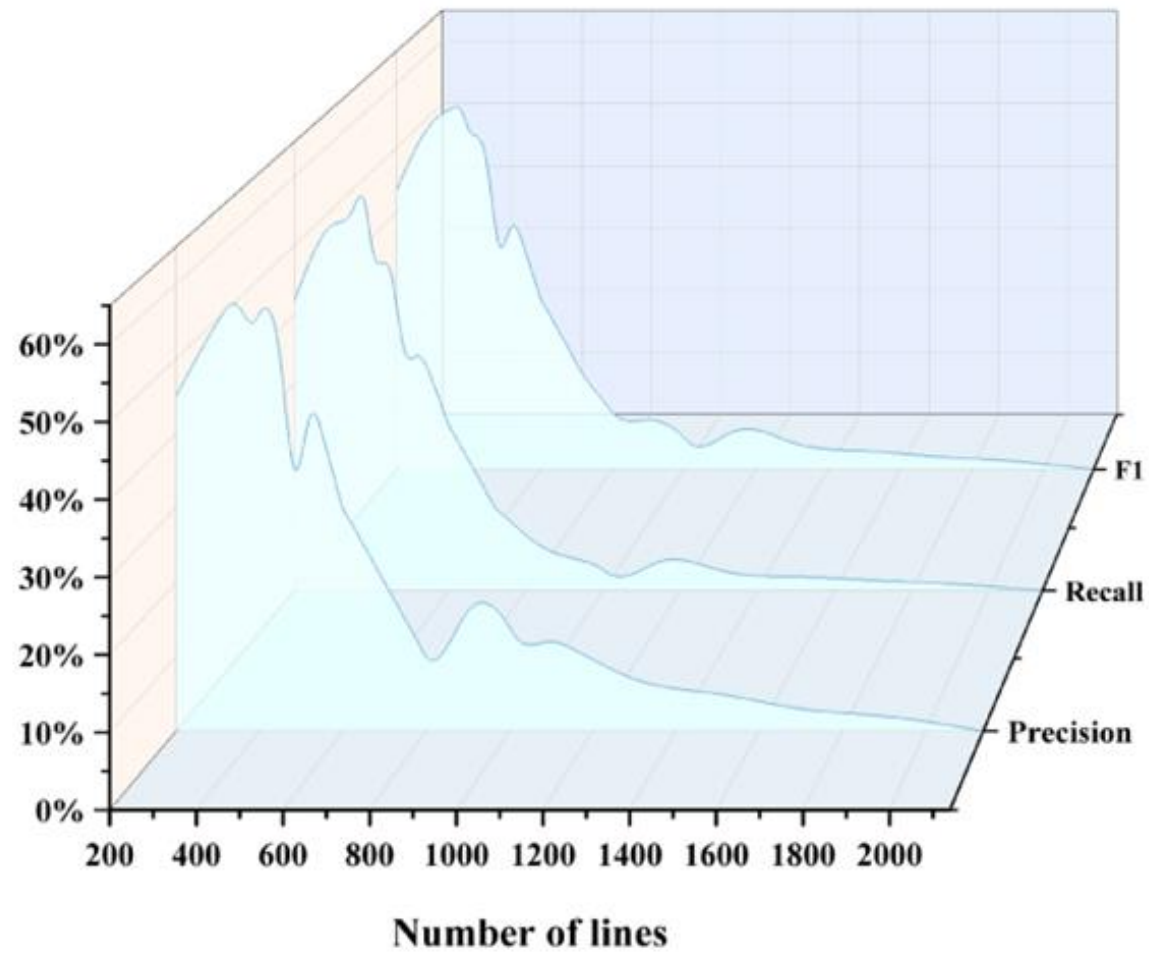

Figure 5: Impact of File Scale (Number of lines) on Evaluation Metrics in End-to-End Pruning

Figure 5 presents the F1 score, precision, and recall curves plotted against LP file size, illustrating the relationship between model performance and problem scale. For small to medium-sized LP files up to approximately 800 lines, the fine-tuned model exhibits meaningful pruning capability despite the limited training data. Performance initially improves with increasing file size, reaching a peak around 350 to 400 lines where the F1 score attains approximately 0.56. This peak indicates that within this context length, the model can effectively capture structural patterns necessary for identifying redundant decision variables.

However, when the LP file size exceeds this optimal range, model performance declines systematically. Beyond approximately 450 lines, all evaluation metrics decrease consistently, indicating simultaneous degradation in both precision and recall. For LP files exceeding 1,500 lines, the F1 score approaches zero, suggesting that the model has largely lost effective pruning capability under the current configuration.

This performance degradation can be attributed to two primary factors. The first factor is that the maximum sequence length used during fine-tuning limits the model's ability to perceive global structure and long-range dependencies in large-scale optimization models. The second factor is that even within its operational context window, the intrinsic complexity of large LP files, which requires joint analysis of dense mathematical expressions, extensive variable interactions, and subtle redundancy patterns, exceeds the pattern recognition capacity established through the small fine-tuning dataset.

Overall, the results demonstrate that LoRA-based fine-tuning with a limited amount of training data is sufficient to achieve effective variable pruning for medium-sized LP models. Extending this capability to large-scale industrial optimization problems will require further methodological advances. In particular, future work should explore pruning strategies that operate in a hierarchical or multi-stage manner to alleviate the complexity of single-pass reasoning, investigate modeling architectures or training paradigms inherently better suited for long-context understanding, and expand the training dataset to include a wider range of large-scale optimization structures. Collectively, these directions are expected to enhance the model's ability to capture global structural dependencies and generalize more robustly to complex, real-world industrial scheduling problems.

## 5. Conclusion

This paper presents EvoOpt-LLM, a large language model–based framework for automating and accelerating key stages of industrial optimization modeling. Unlike prior work that focuses on isolated aspects of model generation or heuristic design, EvoOpt-LLM addresses the full modeling lifecycle through three complementary modules. These modules include automated construction of optimization models from natural-language descriptions, dynamic injection of evolving business constraints into existing MILP models, and end-to-end pruning of redundant decision variables prior to optimization.

Experimental results demonstrate that EvoOpt-LLM can effectively acquire operations research modeling capabilities through parameter-efficient LoRA fine-tuning. In the automated modeling task, the model exhibits strong data efficiency, achieving solver-executable outputs with fewer than 1,500 training samples and reaching a generation rate of 91% and an executability rate of 65.9% with the full dataset. The business constraint injection module reliably augments deployed LP models with solver-compatible constraint blocks while preserving the original objective function and structural integrity, enabling agile iteration of business logic in industrial environments. The variable pruning module further improves computational efficiency by identifying inactive decision variables, achieving meaningful pruning performance on medium-sized LP models despite limited training data.

While the current framework faces challenges in handling very large-scale optimization models due to long-context limitations, the results suggest clear avenues for future improvement. Hierarchical pruning strategies, long-context modeling techniques, and expanded training datasets covering more diverse large-scale structures are expected to further enhance scalability and robustness. Importantly, the proposed approach complements rather than replaces traditional optimization solvers, positioning LLMs as intelligent modeling assistants that operate upstream of numerical optimization.

Overall, EvoOpt-LLM demonstrates that large language models can play a practical and impactful role in industrial operations research, not only by reducing the cost of model construction but also by supporting continuous model evolution and enhancing solver efficiency. This work lays the foundation for future research on tightly integrated LLM–OR frameworks that combine symbolic optimization, data-driven learning, and real-world business logic in a unified decision-support framework.

## Acknowledgement

This work was partially supported by the SEU Kunpeng & Ascend Center of Cultivation.